\documentclass[runningheads]{llncs}
\usepackage{graphicx}
\usepackage{amsmath}
\usepackage{amssymb}
\usepackage{multirow}
\usepackage{float}
\usepackage{booktabs}
\usepackage{makecell}
\usepackage{hyperref}
\begin{document}
%
\title{Unsupervised Learning of Particle Image Velocimetry}
%
%
\author{Mingrui Zhang \and Matthew D. Piggott}
%
\authorrunning{M. Zhang et al.}
%
\institute{Department of Earth Science and Engineering,\\ Imperial College London, South Kensington Campus, London, UK  }

%
\maketitle              
\begin{abstract}
Particle Image Velocimetry (PIV) is a classical flow estimation problem which is widely considered and utilised, especially as a diagnostic tool in experimental fluid dynamics and the remote sensing of environmental flows. Recently, the development of deep learning based methods has inspired new approaches to tackle the PIV problem. These supervised learning based methods are driven by large volumes of data with ground truth training information. However, it is difficult to collect reliable ground truth data in large-scale, real-world scenarios. Although synthetic datasets can be used as alternatives, the gap between the training set-ups and real-world scenarios limits applicability. We present here what we believe to be the first work which takes an unsupervised learning based approach to tackle PIV problems. The proposed approach is inspired by classic optical flow methods. Instead of using ground truth data, we make use of photometric loss between two consecutive image frames, consistency loss in bidirectional flow estimates and spatial smoothness loss to construct the total unsupervised loss function. The approach shows significant potential and advantages for fluid flow estimation. Results presented here demonstrate that our method outputs competitive results compared with classical PIV methods as well as supervised learning based methods for a broad PIV dataset, and even outperforms these existing approaches in some difficult flow cases. Codes and trained models are available at \url{https://github.com/erizmr/UnLiteFlowNet-PIV}.

\end{abstract}
\section{Introduction}
Particle Image Velocimetry (PIV) is one of the most popular measurement techniques in experimental fluid dynamics, and is also used to diagnose flow information from the remote sensing of large-scale environmental flows. The method provides quantitative measurements of velocity fields in fluids that can be used to explore complex flow phenomena. When conducting the PIV technique, the fluid under investigation is seeded with sufficiently small tracer particles (or the presence of naturally occurring features is exploited). These particles are assumed to follow the flow dynamics. With illumination (in the laboratory often through the use of lasers to capture image information over a two-dimensional plane), the particles in the fluid are visible. By comparing resulting flow images between time levels, velocity field information can be inferred \cite{RAdrian}. 
There are two main techniques used for performing classical PIV: cross-correlation and variational optical flow methods. 

The development of deep learning techniques has inspired a new direction for tackling PIV-like problems. Several authors have in the literature proposed and demonstrated the use of supervised learning based methods for PIV. However, due to the unavailability of a broad range of reliable ground truth training data, supervised learning methods have limitations, especially when seeking to generalise to real-world problems. On the other hand, unsupervised learning is a type of machine learning approach that looks for previously undetected patterns in a dataset with no pre-existing labels and with minimum human supervision \cite{Hinton}. 

In this paper we propose a new fluid velocity estimation method using an unsupervised learning strategy based upon particle images.\par

\subsection{Cross-correlation and variational optical flow methods}\label{sect:piv}
There are two main standard approaches for performing particle image velocimetry: cross-correlation and optical flow methods. The cross-correlation method calculates a displacement by searching for the maximum cross-correlation between two interrogation windows from an image pair \cite{Westerweel_cc}, e.g. such as in the {\em WIDIM} (window deformation iterative multi-grid) method. The cross-correlation method is efficient and relatively easy to implement. However, it only outputs a spatially sparse (compared to the resolution of the seed particles in the fluid) displacement field and requires post-processing. The variational optical flow method was proposed by Horn and Schunck (HS) \cite{Horn_optical_flow}. It is a motion estimation approach that has been applied to PIV problems \cite{Ruhnau_of_piv}. It treats the PIV problem through the solution of an optimisation problem, seeking the minimisation of an objective function. The method can output a dense displacement field, but the optimisation process is time-consuming.

\subsection{Deep learning methods}
Machine learning methods, especially deep learning, have made great progress in applications to many real-world problems in recent years. In the PIV community, deep learning has been introduced recently. In \cite{Rabault}, the authors provided a proof-of-concept on this topic, where artificial neural networks are designed to perform end-to-end PIV for the first time in this work. 

PIV techniques are closely related to computational photography, a sub-domain of computer vision. In this community, there are several important works related to the motion estimation problem using deep learning. The {\em FlowNetS} and {\em FlowNetC} networks \cite{Dosovitskiy_flownet} were the first proposed for dense optical flow estimation. {\em FlowNet2} \cite{Ilg_flowne2t}, an extension of {\em FlowNet}, improves the optical flow estimation to a state-of-the-art level. In addition, a lighter-weight network {\em LiteFlowNet} \cite{twhui} has also been proposed. It achieves a similar level accuracy compared to {\em FlowNet2}, using less trainable parameters. Although the networks mentioned above have achieved excellent performance for estimating motion fields from consecutive image pairs, their applications is generally limited to rigid or quasi-rigid motion.

Therefore, it is of interest to explore the performance of these existing networks on particle image velocimetry problems.

\section{Related work}
Supervised and unsupervised learning are two different learning approaches. The key difference is that supervised learning requires ground truth data while unsupervised learning does not. \par

\subsection{Supervised learning methods}\label{sect:sup}
End-to-end supervised learning using neural networks for PIV was first introduced by Rabault et al. in \cite{Rabault}. A convolutional neural network and a fully-connected neural network were trained to perform PIV on several test cases. That work provided a proof-of-concept for the research community. However, the trained model did not achieve the ultimate quality of result compared with traditional PIV methods, and the application scenarios considered were limited to relatively simple cases. Lee et al. \cite{PIV-DCNN} proposed a cascaded network architecture. The network was verified to produce results comparable to standard PIV methods (one-pass cross-correlation method, three-pass window deformation). However, it had larger computational costs and lower efficiency. Another deep architecture approach based on supervised learning was proposed by Cai et al. in \cite{Cai_1}. In that work the author developed a motion estimator {\em PIV-FlowNetS-en} based upon {\em FlowNet}. The estimator is able to extract features from particle images and output dense displacement fields. The model was evaluated both on synthetic and experimental data, and was shown to achieve good accuracy with high efficiency compared to correlation-based PIV methods such as the {\em WIDIM} method. Follow-up work introduced a more complex but lighter-weight network {\em PIV-LiteFlowNet-en} \cite{Cai_2}, based on {\em LiteFlowNet} \cite{twhui}. The model was shown to have the same level of performance as variational optical flow methods in terms of estimation accuracy, while showing advantages in terms of efficiency.

The supervised learning approach relies heavily on large volumes of training data. However, in most real-world scenarios, especially in fluid dynamics, there is no easily available ground truth data and/or it is extremely difficult to annotate the data accurately through human means. Although the use of synthetic data (e.g. based upon computational fluid dynamics studies) can help construct large annotated datasets, the gap between synthetic and real-world scenarios limits the generalisation abilities of the constructed networks. This can mean that supervised learning based approaches may struggle when confronted with data from real-world problems. \par

\subsection{Unsupervised learning methods}
Unsupervised learning is a type of machine learning that, in contrast, looks for previously undetected patterns in a dataset with no pre-existing labels and with minimum human supervision.

To the best of our knowledge, there are no previous examples of approaches that tackle the PIV problem based on unsupervised learning. In the computer vision community, there is some previous work related to the use of unsupervised learning for optical flow estimation. Yu et al. \cite{Yu_back_to}  suggested an unsupervised method based on {\em FlowNet} in order to sidestep the limitations of synthetic datasets. They use a loss function that combines a data term (photometric loss) measuring photometric constancy over time with a spatial term (smoothness loss) modelling the expected variation of flow across the image. In \cite{Meister_unflow}, Meister et al. extended the work using a symmetric, occlusion-aware loss based on both forward and backward optical flow estimates. They also made use of iterative refinement by stacking multiple {\em FlowNet} networks. The model showed advantages and outperformed supervised learning on challenging realistic benchmarks. \par

Our work is inspired in part by Meister et al. \cite{Meister_unflow}; we extend the unsupervised learning strategy to PIV problems, building our model based on {\em LiteFlowNet} instead of {\em FlowNet}. We trained our model on a synthetic PIV dataset generated by Cai et al. in \cite{Cai_1}. Unlike the supervised strategy, we only use the particle images pairs in the dataset, and leave the ground truth motion data (which is used to generate the image pairs) for benchmarking purposes.

\section{Method}
Given a grayscale image pair $I_1,\, I_2: P \rightarrow \mathbb{R}^1$ as input, our goal is to estimate the forward flow field from $I_1$ to $I_2$, ${\bf{F}}^f \equiv (u^f, v^f)^T$, where $u_f$ and $v_f$ are scalar velocity fields in two orthogonal directions. As we take the bidirectional estimate into consideration, the backward flow field is defined as  ${\bf{F}}^b \equiv (u^b, v^b)^T$. In section 3.1 we will introduce the unsupervised loss and how the loss is integrated for training. The network architecture will be described in section 3.2.

\subsection{Unsupervised Loss}
In the training process, the input only contains image pairs ${I_1,\, I_2}$, without the velocity field ground truth. Therefore, we use traditional optical flow measurements to evaluate our results. The total unsupervised loss is a combination of photometric loss, estimate flow smoothness loss and consistency loss between forward and backward fields.

\subsubsection{Photometric loss.}
The photometric loss is defined in terms of the difference between the first image and the warped second image using the forward flow field estimate, and the difference between the second image and the warped first image using the  backward estimate. The bidirectional photometric loss is thus defined as the sum of two parts:
\begin{equation}
    \begin{aligned}
    L_P(I_1, I_2, {\bf{F}}^f, {\bf{F}}^b) = \sum_{{\bf{x}} \in P} &\rho \left(I_1({\bf{x}}) - I_2({\bf{x}}+ {\bf{F}}^f({\bf {x}})) \right) \\
    + &\rho \left(I_2({\bf{x}}) - I_1({\bf{x}} + {\bf{F}}^b({\bf {x}})) \right),
    \end{aligned}
\end{equation}
where $\rho(\cdot) $ is the generalized Charbonnier penalty function, $\rho = (x^2 + \epsilon ^ 2)^\gamma $, which is a differentiable, robust convex function \cite{Sun_IJCV}. We use the values $\gamma = 0.45, \epsilon = 10^{-3}$ in this work.

Image `backwarping' is the key step when computing the photometric loss. In order to make the loss 
backpropagation possible during the training process, we use the differentiable bilinear sampling scheme proposed in \cite{warp}. The basic idea is first to generate a sampling coordinate in target image $I_2$, using $I_1$ and the flow field estimate ${\bf F}^f$. The coordinate can be described as: ${\bf {x_s}} = {{\bf x}}+ {\bf{F}}^f({\bf x}) = (x_1 + F_{u}^f, y_1 + F_{v}^f)$, here $\bf x$ is the coordinate field for image $I_1$. A bilinear sampler is then used to construct the warped image in terms of coordinate $\bf x$:

\begin{equation}
I_{\text{warp}}({\bf {x}}) = \sum_{x_s^i,y_s^i \in {\bf {x_s}}} I_2({\bf{x_s}})\max(0, 1- \left|{x_s}- {x_s^i}\right|)\max(0, 1- \left|{y_s}- {y_s^i}\right|).
\end{equation}

\subsubsection{Smoothness loss.}
\begin{figure}
\begin{center}
\includegraphics[width=0.8\textwidth]{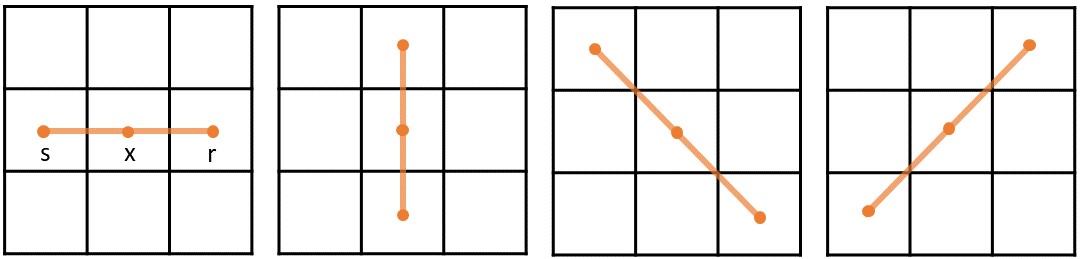}
\caption{Four channel filter used in the smoothness loss: in the directions $x$, $y$ and the two diagonals, shown in the frames abve from left to right respectively. $s,\, x,\, r$ indicate the three neighboring pixels considered for each direction.} \label{filter}
\end{center}
\end{figure}

There are regions in the images that lack necessary information. For example, there may be insufficient particles near image boundaries, as the particles move out of the image area in the second frame or the particles have not entered the image in the first frame. Therefore, to tackle resulting ambiguities, a smoothness loss is included into our total unsupervised loss. To enhance the regularisation effects, we use a second-order smooth constraint \cite{second_order}:

\begin{equation}
    \begin{aligned}
    L_S({\bf{F}}^f, {\bf{F}}^b) = \sum_{({\bf {s}},{\bf {r}}) \in {\bf N(x)}} \sum_{{\bf{x}} \in P} &\rho \left({\bf{F}}^f({\bf s}) - 2{\bf{F}}^f({\bf x}) + {\bf{F}}^f({\bf r})  \right)\\
    + &\rho \left({\bf{F}}^b({\bf s}) - 2{\bf{F}}^b({\bf x}) + {\bf{F}}^b({\bf r})  \right),
    \end{aligned}
\end{equation}
where ${\bf N}$ represents a four channel filter ($x$, $y$ and two diagonals, see Fig. \ref{filter}). Therefore, the process here is first to compute the convolution of the two flow components ($u$ in the $x$ and $v$ in the $y$ directions) with the four channel filter respectively, then compute their Charbonnier loss.

\subsubsection{Consistency loss.}
The forward and backward flow estimates should be consistent, i.e. the forward flow ${\bf F}^f$ is expected to be the inverse of the backward flow ${\bf {F}}^b({\bf x} + {{\bf {F}}^f})$ at the corresponding pixel in the second image. The sum of this pair of flow fields should therefore be zero, and similarly for the backward flow estimate. The consistency loss function can thus be defined as:
\begin{equation}
    \begin{aligned}
    L_C({\bf{F}}^f, {\bf{F}}^b) = \sum_{{\bf{x}} \in P} &\rho \left(  {\bf {F}}^f + {\bf {F}}^b({\bf x} + {{\bf {F}}^f}) \right) \\
    + &\rho \left(  {\bf {F}}^b + {\bf {F}}^f({\bf x} + {{\bf {F}}^b}) \right).
    \end{aligned}
\end{equation}

\subsubsection{Final integrated loss.} The final integrated loss, $L$, combines the above loss terms using weighted (with scalar weights $\lambda_P,\,\lambda_S,\,\lambda_C$)  summation: 
\begin{equation}
    \begin{aligned}
        L(I_1, I_2,{\bf{F}}^f, {\bf{F}}^b) = \lambda_P L_P + \lambda_S L_S + \lambda_C L_C.
    \end{aligned}
    \label{eq:total_loss}
\end{equation}

\subsection{Network architecture}
\subsubsection{UnLiteFlowNet-PIV.} 
Our network, named {\em UnLiteFlowNet-PIV}, is based on {\em LiteFlowNet} \cite{twhui}. It extracts two images' features using a two-stream convolution neural network (NetC) with shared weights. NetC has a pyramidal structure and encodes the image from full resolution to a sixth of that of the original. Then a decoder (NetE) performs cascaded flow inference (convolutionally upsampling) with flow regularisation. The final flow estimate is upsampled to the original resolution using bilinear interpolation. In our work, we compute both the forward and backward flow in one estimation. The input for the forward flow estimation is $(I_1, I_2)$, and it is $(I_2, I_1)$ for the backward flow. 

\begin{figure}
\includegraphics[width=\textwidth]{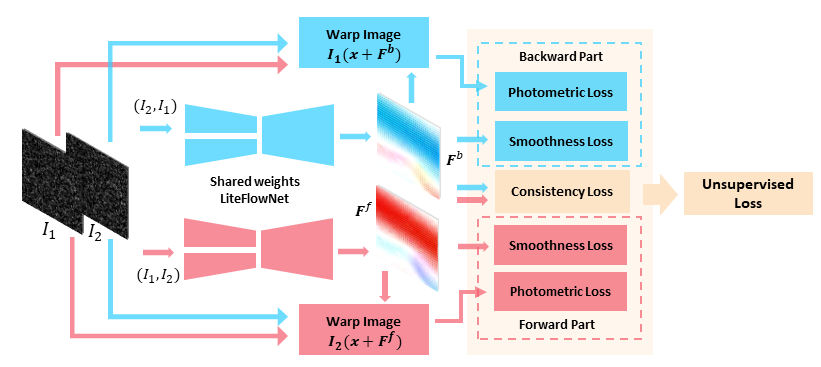}
\caption{Data flow for {\em UnLiteFlowNet-PIV}, from inputs to unsupervised loss. Due to taking bidirectional flows into consideration, the red components represent the forward part, with the image pair $(I_1, I_2)$ as input. The blue components indicate the backward part. Although there are two networks shown for clarity, since shared weights are used there is only one in the implementation.} \label{network}
\end{figure}

\subsubsection{Training Loss.}
The training loss function's design is similar to that in {\em FlowNet} \cite{Dosovitskiy_flownet} and {\em LiteFlowNet} \cite{twhui}, and uses a multi-scale resolution loss. It is the weighted sum of the estimation losses from each of the intermediate layers:

\begin{equation}
    \begin{aligned}
        L_T = \sum_{i} w_i L_i,
    \end{aligned}
\end{equation}
where $L_i$ is the loss function \eqref{eq:total_loss}. At each layer, the image pair $(I_1, I_2)$ is downsampled to compute the current layer's loss. As the distance between pixels effectively changes after downsampling, the flow estimate is multiplied by the appropriate scaling factor, which is the fraction between the current and the full image resolution. Here, $i_{\text{max}} = 6$, and $L_6$ indicates the loss at full resolution.

\section{Evaluation}
\subsection{PIV dataset}
The dataset considered in this work was generated by Cai et al. \cite{Cai_1}. The dataset contains 15,050 particle image pairs with the originating flow field ground truth data obtained from computational fluid dynamics simulations. There are eight different types of flow contained in the dataset, including `uniform' flow, flow past a backward facing step (`back-step') and past a `cylinder', both at a variety of Reynolds numbers, `DNS-turbulence', sea surface flow driven by a quasi-geostrophic model (`SQG'), etc. Detailed information on the dataset is provided in Table \ref{table:piv_dataset}. In our work we use half of the dataset for training and the other half for testing.

\subsection{Training details}
We train the model for 40,000 iterations with a batch size of four image pairs using the Adam optimiser. The learning rate is kept at $10^{-4}$. The smoothness loss weight is $\lambda_C = 3.0$, the consistency loss weight $\lambda_C = 0.2$, photometric loss weight $\lambda_P = 1.0$. The weights for different layers are set to [12.7, 5.5, 4.35, 3.9, 3.4, 1.1] as in \cite{Meister_unflow}, from the full resolution to the lowest level. The image pair are normalized from value ranges of 0--255 to 0--1 before feeding into the network.

\begin{figure}
\includegraphics[width=\textwidth]{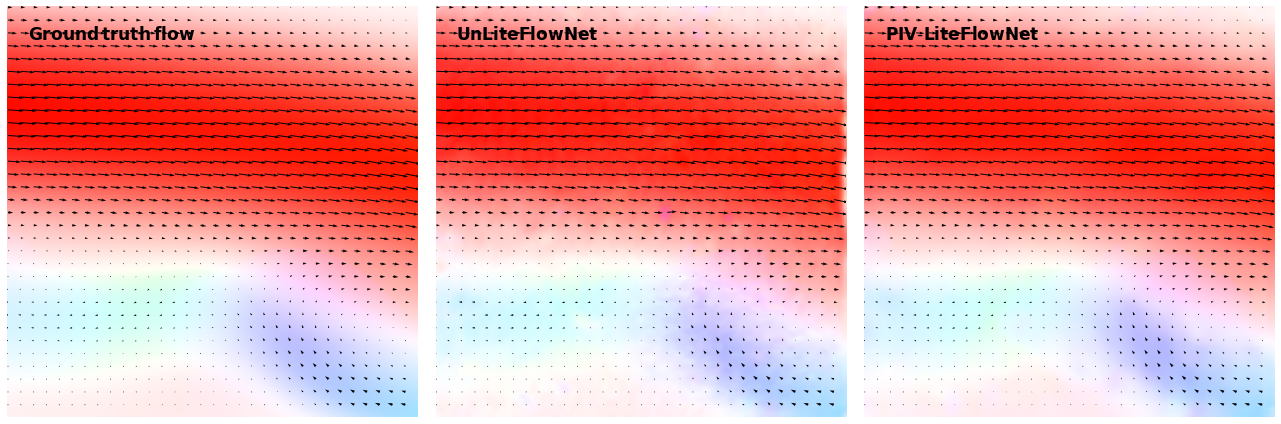}
\includegraphics[width=\textwidth]{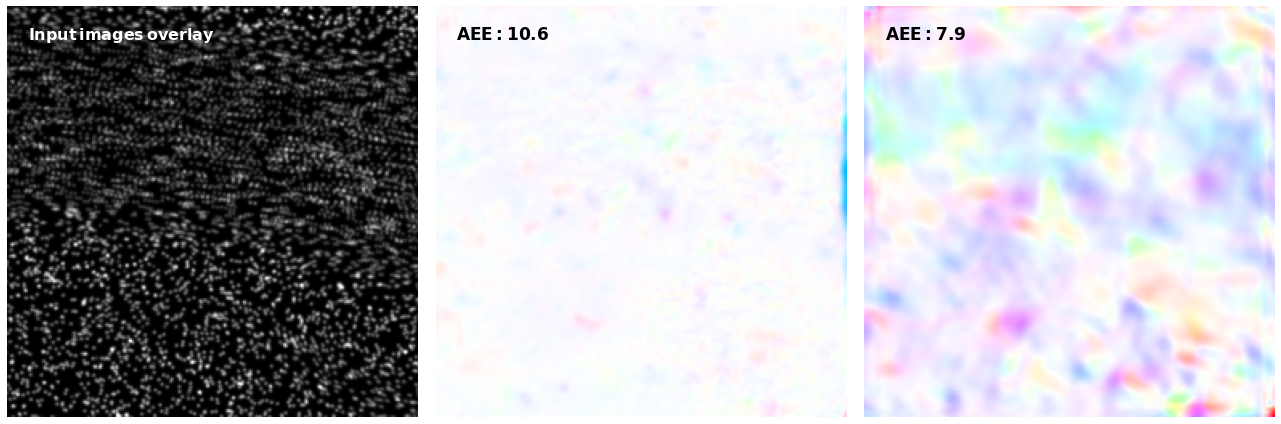}
\includegraphics[width=\textwidth]{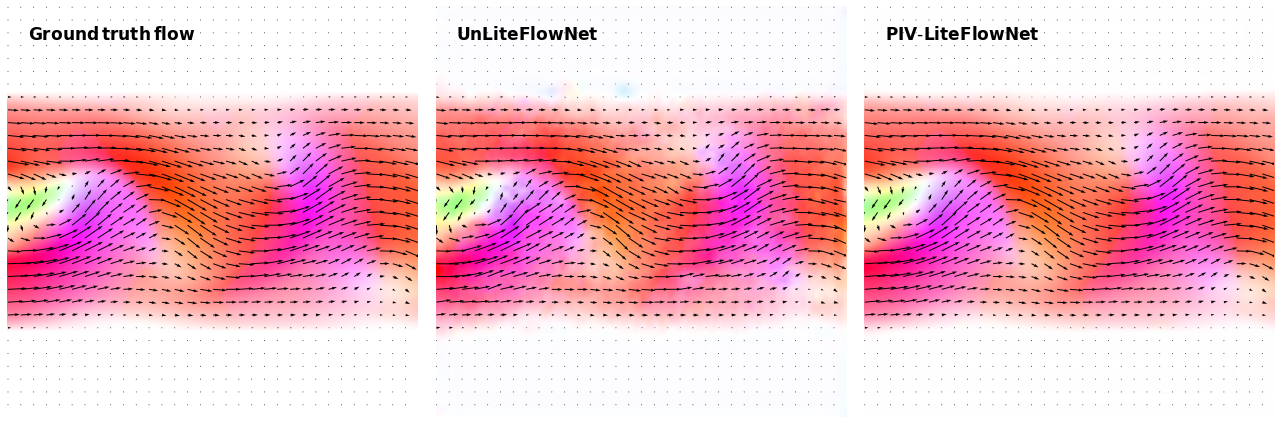}
\includegraphics[width=\textwidth]{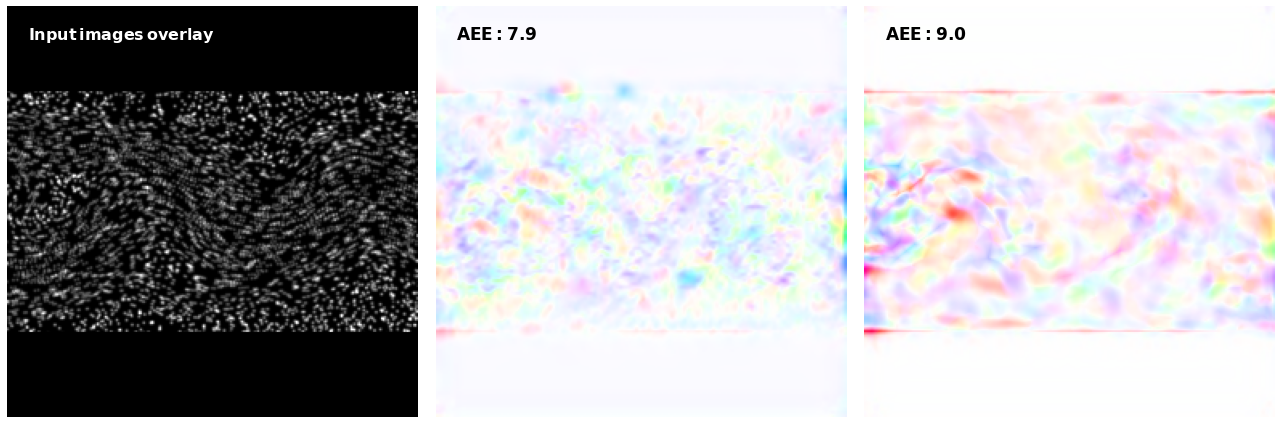}
\caption{Visual comparisons between ground truth flow data (left), our fully unsupervised model {\em UnLiteFlowNet-PIV} (middle) and {\em PIV-LiteFlowNet} (right) shown on the first and third rows for respectively the `Back-step' and `Cylinder' flow cases. The flow field colour is coded in HSV \cite{HSV}. The second and fourth rows show the input images overlays (left), along with the errors and corresponding Averaged Endpoint Error (AEE) values for the two networks and image pairs considered. In the error plots the white colour indicates zero error, and pixel colour with higher saturation represents larger errors. It can be observed that even though our new unsupervised model never has access to the ground truth during training, it still tends to outperform the supervised model. See also Figs. \ref{results_2} and \ref{results_3} for further cases.} \label{results_1}
\end{figure}

\begin{figure}
\includegraphics[width=\textwidth]{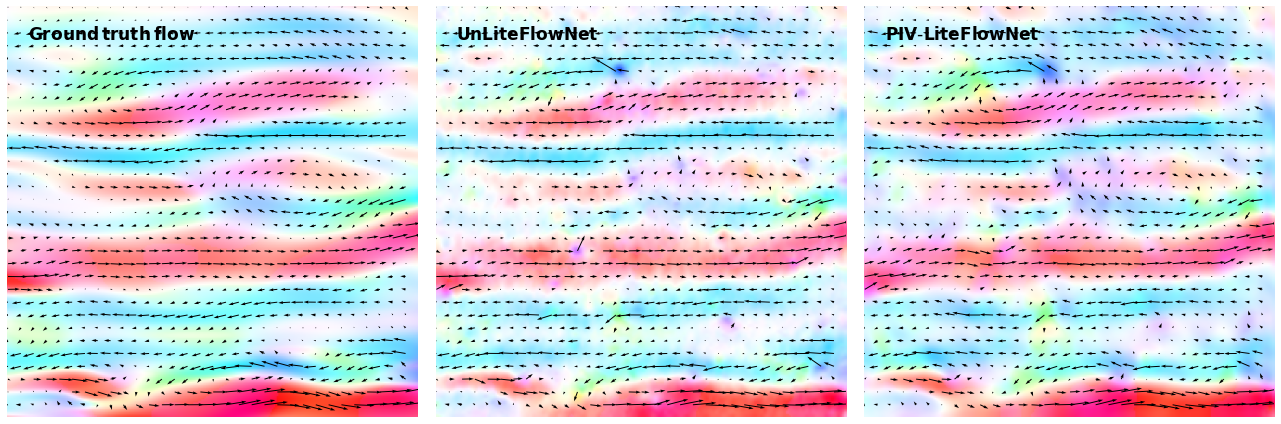}
\includegraphics[width=\textwidth]{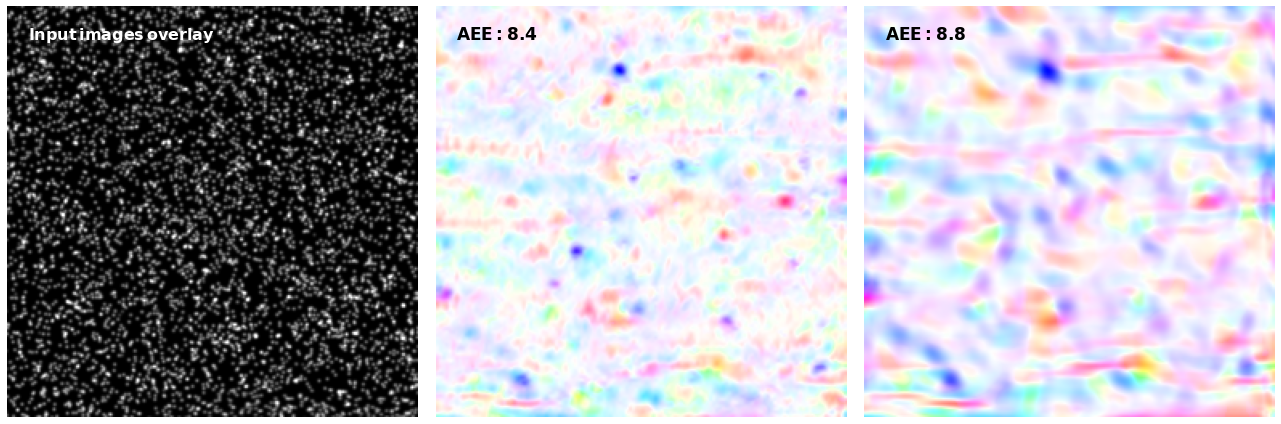}
\includegraphics[width=\textwidth]{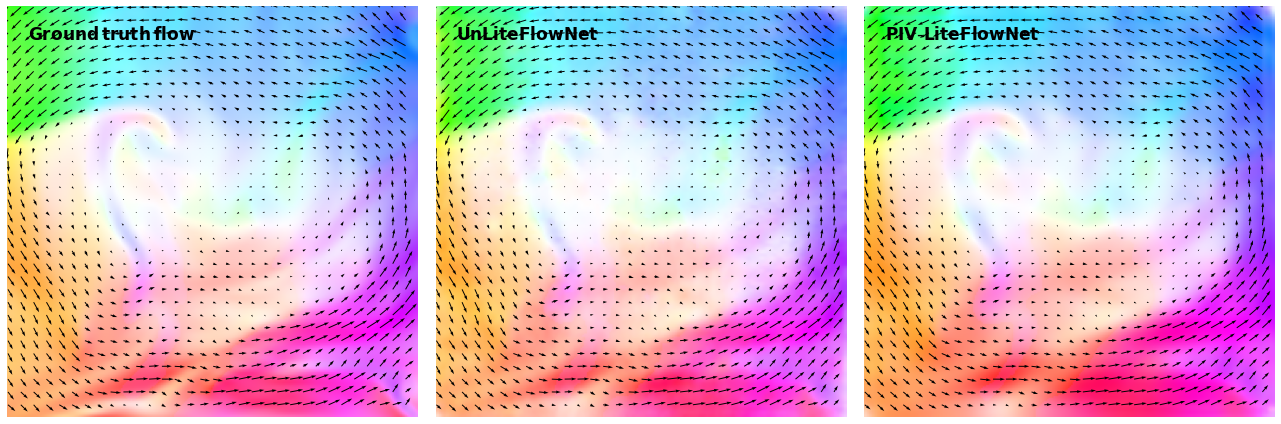}
\includegraphics[width=\textwidth]{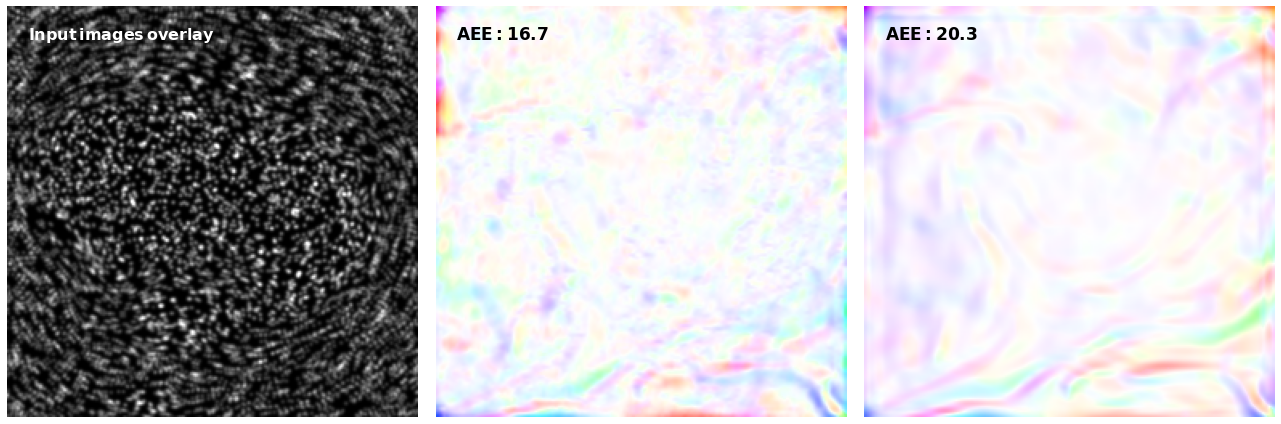}

\caption{Caption as for Fig. \ref{results_1} but now for the flow cases `JHTDB-channel' (first two rows) and `DNS-turbulence' (final two rows).
It can be observed that errors are distributed along the vortex boundaries in both models. However, our {\em UnLiteFlowNet-PIV} model does tend to demonstrate lower errors in the interior, e.g. around the regions with strong internal vortices.} \label{results_2}
\end{figure}

\subsection{Results}
Table \ref{table:results} compares the accuracy of our model with previous work and different approaches, including classical PIV and state-of-the-art deep learning based methods. The results are evaluated on the PIV dataset, with the Averaged Endpoint Error (AEE) calculated for different flow types. In order to compare the results easier, we set the units of the AEE to pixel per 100 pixels. The AEE can be described as the $L^2$-norm of the difference in flow estimation ${\bf{F}}^e$ and the flow ground truth ${\bf{F}}^g$:

\begin{equation}
    {\text{AEE}} = \left\Vert {\bf{F}}^e -{\bf{F}}^g\right\Vert_2.
\end{equation}

\subsubsection{Comparison to classical PIV.} 
It can be observed that our unsupervised model outperforms classical correlation-based PIV WIDIM methods in almost all flow cases, especially for the challenging cases of DNS-turbulence and SQG. Although the unsupervised model does not outperform the Horn–Schunck (HS) optical flow method \cite{Horn_optical_flow}, the differences are relatively small. In addition, as mentioned above, the HS optical flow method requires a large amount of computational time in order to conduct the optimisation process, which results in low efficiency especially when multiple image pairs need to be processed. Without considering the time to load images from disk, the computational time for 500 image (256 $\times$ 256) pairs using our {\em UnLiteFlowNet-PIV} is 10.17 seconds on an Nvidia Tesla P100 GPU, while the HS optical method requires roughly 556.5 seconds and WIDIM (with a window size of 29 $\times$ 29) requires 211.5 seconds on an Intel Core I7-7700 CPU \cite{Cai_2}. Although the classical PIV methods are tested on a CPU, as shown in the \cite{Rabault,Cai_1} the speed improvements for them running on GPUs are limited. Therefore, efficiency is a great advantage for learning based methods compared to the classical approaches.

\begin{figure}
\includegraphics[width=\textwidth]{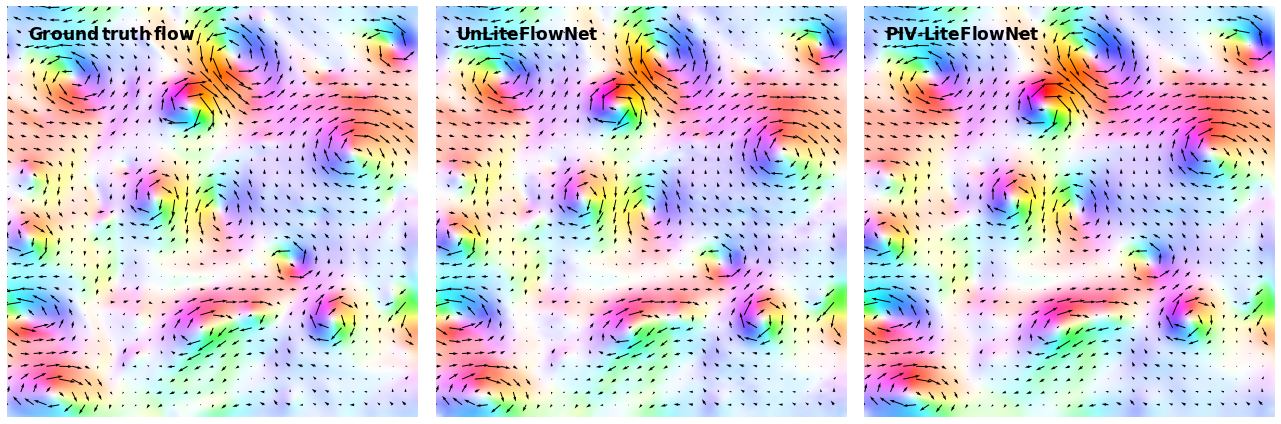}
\includegraphics[width=\textwidth]{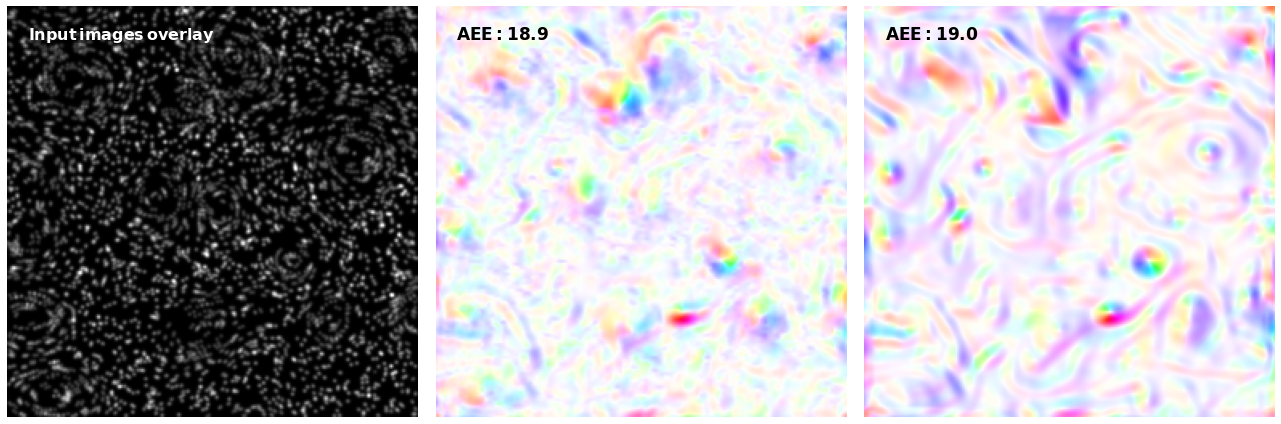}
\caption{Caption similar to Fig. \ref{results_1}, but now only showing the `SQG' case. The {\em UnLiteFlowNet-PIV} model again shows more than competitive performance on this challenging flow case.} \label{results_3}
\end{figure}

\subsubsection{Comparison to deep learning PIV.} 
The unsupervised learning approach shows potentially significant advantages compared to state-of-the-art supervised learning methods. Figs. \ref{results_1}, \ref{results_2} and \ref{results_3} demonstrate comparisons between our fully unsupervised model {\em UnLiteFlowNet-PIV} and {\em PIV-LiteFlowNet}. {\em PIV-LiteFlowNet} \cite{Cai_2} uses a similar network architecture to our {\em UnLiteFlowNet-PIV}, but is trained using a supervised learning strategy with ground truth data. Although the unsupervised {\em UnLiteFlowNet-PIV} never has access to the ground truth data during the training process, it still outperforms most supervised learning methods ({\em PIV-NetS-noRef}, {\em PIV-NetS}, {\em PIV-LiteFlowNet}), especially on difficult cases. Therefore, the unsupervised learning method with an accurate loss function shows competitive capabilities and often better performance compared to supervised methods. 

{\em PIV-LiteFlowNet-en} \cite{Cai_2} is an enhanced version of {\em PIV-LiteFlowNet}, it adds one additional layer at the end of the NetE, which improves its inference ability but makes the network more complicate and heavier. We did not try to construct deeper networks in our work for brevity. There are ideas for improving the performance by stacking networks \cite{Ilg_flowne2t}, which would also be an interesting avenue to explore in further work.

\begin{table}
\caption{Averaged Endpoint Error (AEE) for the PIV dataset (averaged over all impage pairs), the error unit is set to pixel per 100 pixels for easier comparison. From top to bottom, WIDIM and HS optical flow are the classical PIV methods described in section \ref{sect:piv}, the next four rows are state-of-the-art supervised learning methods described in section \ref{sect:sup}. The final row shows results of our unsupervised method introduced in this work.}\label{table:results}

\begin{tabular}{@{\extracolsep{4pt}}c|cccccccccc@{}}
\toprule
\multirow{3}{*}{Methods} & 
\multicolumn{2}{c}{\multirow{2}{*}{Back-Step}}&
\multicolumn{2}{c}{\multirow{2}{*}{Cylinder}}&  
\multicolumn{2}{c}{JHTDB}&
\multicolumn{2}{c}{DNS}&
\multicolumn{2}{c}{\multirow{2}{*}{SQG}}\\ 
& & & & &\multicolumn{2}{c}{channel} &\multicolumn{2}{c}{turbulence}  && \\
\cline{2-3} \cline{4-5} \cline{6-7} \cline{8-9}\cline{10-11}
& train& test& train&test &train&test  &train&test &train&test\\
\midrule
WIDIM \cite{Cai_2}&- & 3.4 & - & 8.3 & -& 8.4 & -& 30.4&-&45.7  \\
HS optical flow \cite{Cai_2}&- & 4.5  & -& 7.0 & -& 6.9 & -& 13.5 &-& 15.6  \\
\midrule
PIV-NetS-noRef \cite{Cai_1} & 13.6 & 13.9  & 19.8 & 19.4& 24.6& 24.7 & 50.6& 52.5& 51.9 & 52.5 \\
PIV-NetS \cite{Cai_1} & 5.8& 5.9  & 6.9 & 7.2 &16.3& 15.5 & 27.1& 28.2 &28.9& 29.4      \\ 
\midrule
PIV-LiteFlowNet \cite{Cai_2}& 5.5 & 5.6   & 8.7  & 8.3 &  10.9 & 10.4  & 18.8 & 19.6  & 19.8 & 20.2   \\
PIV-LiteFlowNet-en \cite{Cai_2}&3.2 & 3.3 & 5.2  & 4.9  & 7.9 & 7.5 &11.6 & 12.2 & 12.4 & 12.6  \\
\midrule
\multirow{2}{*}{\bf UnLiteFlowNet-PIV}& \multirow{2}{*}{-} & \multirow{2}{*}{10.1} &  \multirow{2}{*}{-}   & \multirow{2}{*}{7.8} & \multirow{2}{*}{-} &  \multirow{2}{*}{9.6} & \multirow{2}{*}{-} & \multirow{2}{*}{13.5} & \multirow{2}{*}{-} & \multirow{2}{*}{19.7}   \\
& &  &    &  &  &   & &  &  &    \\
\bottomrule
\end{tabular}
\end{table}

\begin{table}
\centering
\caption{Averaged Endpoint Error (AEE) on test dataset for models trained by different loss functions. The error unit is set to pixel per 100 pixels for easier comparison.}\label{table:ablation study}
\begin{tabular}{@{\extracolsep{4pt}}c|ccccc@{}}
\toprule
\multirow{2}{*}{Loss function} & 
\multicolumn{1}{c}{\multirow{2}{*}{Back-Step}}&
\multicolumn{1}{c}{\multirow{2}{*}{Cylinder}}&  
\multicolumn{1}{c}{JHTDB}&
\multicolumn{1}{c}{DNS}&
\multicolumn{1}{c}{\multirow{2}{*}{SQG}}\\ 
  & & &\multicolumn{1}{c}{channel} &\multicolumn{1}{c}{turbulence}  &  \\
\midrule
$L_P+L_S+L_C$ &  10.1  & 7.8 & 9.6  & 13.5 & 19.7\\
\midrule
$L_P+L_S$ &  11.6  & 10.5 & 15.3  & 21.4 & 22.5\\
\midrule
$L_P+L_C$ &  14.1  & 38.4 & 18.1  & 23.6 & 25.5\\
\bottomrule
\end{tabular}
\end{table}

\subsubsection{Ablation study.} There are three components to the loss function as mentioned above. The contributions to model performance of each component are investigated here. Results are summarize in table \ref{table:ablation study}. The model is trained for 40,000 iterations with three different loss functions: $L_P+L_S+L_C$ (i.e. the full loss function), $L_P+L_S$ (no consistency loss), and $L_P+L_C$ (no smoothness loss). The model trained using the full loss performs the best among the three on the test dataset. Removing either smoothness or consistency loss leads to a worse performance on the test dataset considered here.

\section{Conclusion}
We present here the first work using an unsupervised learning approach for solving Particle Image Velocimetry (PIV) problems. The proposed unsupervised learning approach shows significant promise and potential advantages for fluid flow estimation. It yields competitive results compared with classical PIV methods as well as existing supervised learning based methods, and even outperforms them on some difficult flow cases. 
Furthermore, the unsupervised learning method does not rely on any ground truth data in order to train, which makes it extremely promising to generalize to complex real-world flow scenarios where ground truth is effectively unknowable, and thus represents a key advantage over supervised methods.

\begin{table}
\caption{Detailed description of the PIV dataset considered, from \cite{Cai_1}. $dx$ refers to the particle displacements considered between two image frames in units of number of pixels. Re refers to the Reynolds numbers considered. `JHTDB' implies that the data was taken from the Johns Hopkins turbulence databases \cite{JHTDB}. Refer to \cite{Cai_1} for further details.}\label{table:piv_dataset}
\begin{center}
\begin{tabular}{@{\extracolsep{4pt}}cccc@{}}
\toprule
Metric Name& Description & Condition& Quantity\\
\toprule
Uniform & Uniform flow & $|dx| \in [0, 5]$ & 1000   \\
\midrule
\multirow{4}{*}{Back-step}&\multirow{4}{*}{Flow past a backward facing step} 
   & Re = 800  & 600  \\
 &  & Re = 1000  & 600  \\
 &  & Re = 1200  & 1000 \\ 
 &  & Re = 1500  & 1000 \\  
\midrule
\multirow{5}{*}{Cylinder}&\multirow{5}{*}{Flow past a circular cylinder} & Re = 40  & 50  \\
 &  & Re = 150  & 500  \\
 &  & Re = 200  & 500 \\ 
 &  & Re = 300  & 500 \\
 &  & Re = 400  & 500 \\
\midrule
\multirow{2}{*}{DNS-turbulence} &  Homogeneous and    & \multirow{2}{*}{-}  & \multirow{2}{*}{2000}  \\
 &isotropic turbulent flow &  &  \\
\midrule
\multirow{2}{*}{SQG} & Sea surface flow     & \multirow{2}{*}{-}  & \multirow{2}{*}{1500}  \\
 &driven by SQG model &  &  \\
\midrule
\multirow{2}{*}{Channel flow } &  Channel flow    & \multirow{2}{*}{-}  & \multirow{2}{*}{1600}  \\
 &provided by JHTDB &  &  \\
\midrule
\multirow{2}{*}{JHTDB-mhd1024 } &  Forced MHD turbulence     & \multirow{2}{*}{-}  & \multirow{2}{*}{800}  \\
 &provided by JHTDB  &  &  \\
\midrule
\multirow{2}{*}{JHTDB-isotropic1024  } &  Forced isotropic turbulence   & \multirow{2}{*}{-}  & \multirow{2}{*}{2000}  \\
 &provided by JHTDB &  &  \\
\bottomrule
\end{tabular}
\end{center}
\end{table}

\subsection*{Acknowledgements}
The authors would like to acknowledge funding from the Chinese Scholarship Council and Imperial College London
(a pump priming research award from the Energy Futures Lab, Data Science Institute and Gratham Institute -- Climate Change and the Environment) that supported this work.

%
%

%
%
%
\bibliographystyle{splncs04}
\bibliography{mybibliography}

\end{document}